\documentclass[conference]{IEEEtran}
\IEEEoverridecommandlockouts
\usepackage{cite}
\usepackage{amsmath,amssymb,amsfonts}
\usepackage{algorithmic}
\usepackage{graphicx}
\usepackage{textcomp}
\usepackage{xcolor}
\def\BibTeX{{\rm B\kern-.05em{\sc i\kern-.025em b}\kern-.08em
    T\kern-.1667em\lower.7ex\hbox{E}\kern-.125emX}}

\makeatletter % changes the catcode of @ to 11
\newcommand{\linebreakand}{%
  \end{@IEEEauthorhalign}
  \hfill\mbox{}\par
  \mbox{}\hfill\begin{@IEEEauthorhalign}
}
\makeatother % changes the catcode of @ back to 12
\begin{document}

\title{Improving Academic Skills Assessment with NLP and Ensemble Learning\\
}

\author{
\IEEEauthorblockN{Xinyi Huang}
\IEEEauthorblockA{\textit{University of Chicago} \\
Chicago, USA \\
bellaxinyihuang@gmail.com}
\and
\IEEEauthorblockN{Yingyi Wu}
\IEEEauthorblockA{\textit{Rensselaer Polytechnic Institute} \\
Seattle,  USA \\
 wuyingyi1104@gmail.com}
\and
\IEEEauthorblockN{Danyang Zhang}
\IEEEauthorblockA{\textit{san jose state university } \\
San Jose, USA \\
josephzdy@gmail.com}
\linebreakand
\and
\IEEEauthorblockN{Jiacheng Hu}
\IEEEauthorblockA{\textit{Tulane University } \\
New Orleans, USA \\
jhu10@tulane.edu}
\and
\IEEEauthorblockN{Yujian Long}
\IEEEauthorblockA{\textit{Independent Researcher} \\
 Frisco, TX, USA \\
longyujian@gmail.com}
}

\maketitle

\begin{abstract}
This study addresses the critical challenges of assessing foundational academic skills by leveraging advancements in natural language processing (NLP). Traditional assessment methods often struggle to provide timely and comprehensive feedback on key cognitive and linguistic aspects, such as coherence, syntax, and analytical reasoning. Our approach integrates multiple state-of-the-art NLP models, including BERT, RoBERTa, BART, DeBERTa, and T5, within an ensemble learning framework. These models are combined through stacking techniques using LightGBM and Ridge regression to enhance predictive accuracy. The methodology involves detailed data preprocessing, feature extraction, and pseudo-label learning to optimize model performance. By incorporating sophisticated NLP techniques and ensemble learning, this study significantly improves the accuracy and efficiency of assessments, offering a robust solution that surpasses traditional methods and opens new avenues for educational technology research focused on enhancing core academic competencies.
\end{abstract}


\section{Introduction}
The assessment of English Language Learners (ELL) in grades 8-12 presents significant challenges, particularly in evaluating cohesion, syntax, vocabulary, phraseology, grammar, and conventions. Traditional methods often fail to provide timely and comprehensive feedback necessary for student improvement and instructional support. This study leverages recent advancements in natural language processing (NLP) to develop a robust model that enhances the accuracy and efficiency of these assessments.

Central to our approach is the integration of multiple state-of-the-art NLP models within an ensemble learning framework. BERT, introduced by Devlin et al. (2019), revolutionized text analysis by capturing context bidirectionally. Building on BERT's success, Liu et al. (2019) developed RoBERTa, which further optimized training procedures, resulting in improved performance across various tasks. Similarly, Lewis et al. (2020) introduced BART, combining bidirectional and autoregressive Transformers to enhance text generation and comprehension capabilities.

To address the specific needs of educational assessment, our model incorporates these advanced NLP techniques along with DeBERTa, which employs disentangled attention mechanisms to capture nuanced textual dependencies. Additionally, T5’s text-to-text framework, as explored by Raffel et al. (2020), allows flexible task handling by converting all NLP tasks into a text-to-text format. These models are integrated through stacking, a technique where multiple base models' predictions are combined using meta-learners like LightGBM and Ridge regression. LightGBM, known for its efficiency in handling large-scale data through gradient boosting, and Ridge regression, which provides regularization to ensure stable predictions, are crucial in achieving high predictive accuracy.

Our methodology begins with comprehensive data preprocessing. Essays are processed using multi-label stratified cross-validation to maintain balanced representation across all linguistic indicators. Text data is tokenized using pre-trained tokenizers, ensuring consistency and maximizing model performance. We employ custom PyTorch model classes, such as MeanPooling and DebertaBaseModel, to handle text inputs and perform classification tasks effectively.

Feature extraction and pseudo-label learning play significant roles in refining our model's performance. By extracting features from the last four layers of 38 pre-trained models and employing forward feature selection, we identify optimal configurations for Support Vector Regression (SVR). Pseudo-label learning involves using both pre-trained and newly generated pseudo-labeled data to fine-tune DeBERTa models, enhancing their generalization capabilities across diverse datasets.

In conclusion, our ensemble learning approach integrates advanced NLP models like DeBERTa, RoBERTa, T5, and GPT through sophisticated stacking techniques. This method, combined with robust data preprocessing, feature extraction, and pseudo-label learning strategies, significantly improves the accuracy of linguistic assessments for ELL students. This study not only addresses the limitations of traditional assessment methods but also sets the stage for future research in applying advanced ensemble learning techniques to educational domains.

\section{Related Work}
Recent advancements in natural language processing (NLP) have significantly improved text analysis and understanding, providing new opportunities for educational assessments. Devlin et al.\cite{devlin2018bert} introduced BERT, which transformed the landscape of language representation models through bidirectional training of Transformer encoders. BERT's ability to capture context from both directions in a text significantly improved performance on various NLP tasks, including text classification and language inference. Building on this, Liu et al.\cite{liu2019roberta} developed RoBERTa, optimizing the training process by using more data and larger batches, leading to further improvements in model performance. Similarly, Lewis et al.\cite{lewis2019bart} introduced BART, which combines bidirectional and autoregressive Transformers, enhancing the model's ability to generate and comprehend text.

While these models achieved state-of-the-art results in many benchmarks, their application to educational assessments, particularly for predicting multiple linguistic indicators simultaneously, remained underexplored. Sun et al.\cite{sun2019fine} demonstrated the potential of fine-tuning BERT for specific tasks, highlighting its adaptability. Raffel et al.\cite{raffel2020exploring} extended this further with the introduction of T5, a text-to-text transfer learning framework, showcasing the versatility of transfer learning in handling various NLP tasks.

Ensemble learning approaches, such as those discussed by Sagi and Rokach\cite{sagi2018ensemble} , have shown promise in combining the strengths of individual models to improve overall performance. Techniques like stacking and blending have been effective in various domains. Krawczyk et al.\cite{krawczyk2017ensemble} provided a comprehensive survey on ensemble learning for data stream analysis, emphasizing its robustness and adaptability.

Yang et al.\cite{yang2019xlnet} introduced XLNet, which further advanced the capabilities of autoregressive pretraining for language understanding, showcasing improvements over BERT in several benchmarks. Howard and Ruder \cite{howard2018universal} proposed universal language model fine-tuning for text classification, demonstrating significant performance gains in various text classification tasks. Qiu et al.\cite{qiu2020pre} provided a comprehensive survey on pre-trained models for NLP, emphasizing their impact on various downstream tasks.

Brown et al.\cite{brown2020language} introduced a groundbreaking model that showcased the ability of language models to perform well with few-shot learning, further emphasizing the potential of pre-trained models in NLP. Conneau and Lample\cite{conneau2019cross} discussed cross-lingual language model pretraining, highlighting the benefits of multilingual pretraining for cross-lingual transfer tasks. Williams et al.\cite{williams2017broad} presented a broad-coverage challenge corpus for sentence understanding through inference, which has been widely used to benchmark NLP models.

He et al.\cite{he2024utilizing} introduce methods for utilizing large language models to identify constraints, which informs our approach to optimizing data preprocessing and feature extraction with NLP models like BERT and RoBERTa in educational assessments.

Sun and Ortiz\cite{sun2024ai}provide an AI-based system that uses LLMs for complex activity tracking, which parallels our use of multiple NLP models and pseudo-label learning to improve coherence and accuracy in assessments.

Yu et al.\cite{yu2024enhancing} study large language models for medical question answering, highlighting techniques with BART and T5 that enhance our strategy for generating contextually relevant and accurate feedback.

Zhang et al.\cite{zhang2024fairness} explore fairness-aware feature selection using causal graphs, supporting our use of LightGBM and Ridge regression to maintain fairness and reduce bias in model ensemble learning.

Radford et al.\cite{radford2018improving} introduced generative pre-training of language models, which has had a significant impact on subsequent NLP research. The GLUE benchmark proposed by Wang et al.\cite{wang2018glue} has been instrumental in evaluating the performance of NLP models across various tasks, providing a standardized framework for comparison.

Our research builds on these advancements by proposing an ensemble method that integrates multiple pre-trained models, fine-tunes them for the specific task of linguistic assessment, and combines their outputs using LightGBM and Ridge regression. This approach not only enhances prediction accuracy but also provides a scalable and efficient solution for real-world educational applications.

\section{Methodology}
\begin{figure*}[h]
\centering
\includegraphics[width=1.0\textwidth]{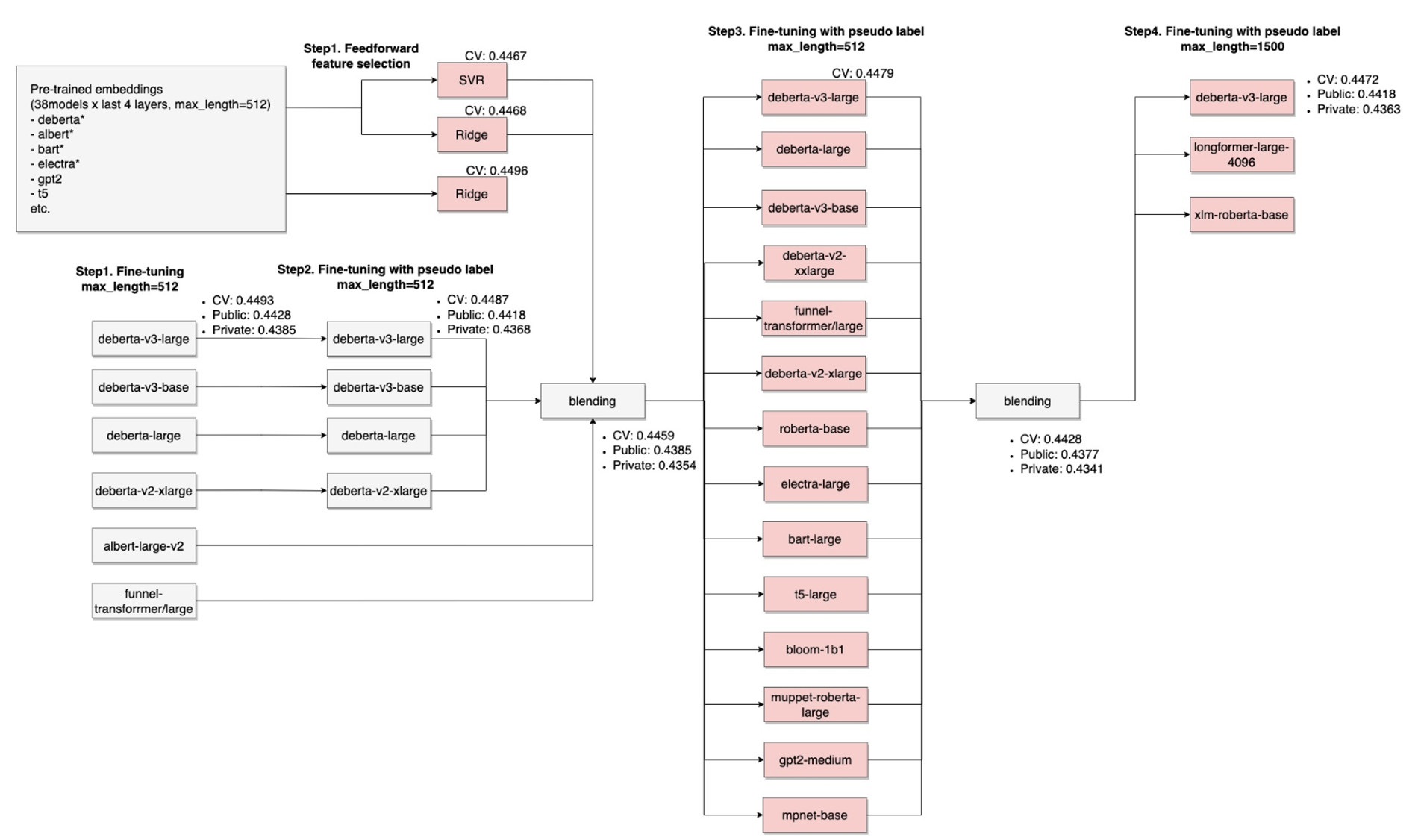}
\caption{Model ensemble structure for organ models}
\label{fig:ensemble_structure}
\end{figure*}

Multi-label text classification is a challenging task due to the interdependencies between labels and the variability in text length and structure. In this section, we employ a series of sophisticated techniques to preprocess data, design model architectures, and evaluate performance. This paper presents an advanced approach for multi-label text classification using a combination of stratified cross-validation, pseudo-labeling, and model stacking. The methodology leverages a diverse set of pre-trained models and integrates their predictions using ensemble techniques to achieve robust performance. The whole model pipeline is shown in Fig \ref{fig:ensemble_structure}

\subsection{Feature extraction}
we trained SVR/Ridge using the pre-trained model embeddings, extracted features from the last 4 layers of 38 pre-trained models, and used forward feature selection to explore the best SVR, and trained the Ridge model using the best embedding combination of SVR, which was my best single model with a CV of 0.4467. And fed the features as maks input.
\subsubsection{SVR}
Support Vector Regression (SVR) is a type of Support Vector Machine (SVM) that is used for regression tasks. It attempts to fit the best line within a predefined margin of tolerance, $\epsilon$. The SVR model is defined by:
\begin{equation}
\min_{\mathbf{w}, b} \frac{1}{2} \|\mathbf{w}\|^2 + C \sum_{i=1}^{N} \max(0, |y_i - (\mathbf{w} \cdot \mathbf{x}_i + b)| - \epsilon)
\end{equation}
where $\mathbf{w}$ is the weight vector, $b$ is the bias term, $C$ is the regularization parameter, $y_i$ is the true value, and $\mathbf{x}_i$ is the input feature vector. SVR aims to find a function that approximates the true relationship between the features and the target variable while minimizing prediction errors within the $\epsilon$ margin.

\subsubsection{Ridge Regression}
Ridge Regression, also known as Tikhonov regularization, is a linear regression model that includes a regularization term to prevent overfitting. The Ridge regression model solves the following optimization problem:
\begin{equation}
\min_{\mathbf{w}} \|\mathbf{y} - \mathbf{X} \mathbf{w}\|^2 + \lambda \|\mathbf{w}\|^2
\end{equation}
where $\mathbf{y}$ is the vector of observed values, $\mathbf{X}$ is the matrix of input features, $\mathbf{w}$ is the weight vector, and $\lambda$ is the regularization parameter. The regularization term $\lambda \|\mathbf{w}\|^2$ penalizes large weights, encouraging the model to find a balance between fitting the training data and maintaining simplicity in the model.

\subsection{Model Class}
The architecture consists of two primary PyTorch models: MeanPooling and CustomModel. The MeanPooling class performs mean pooling on hidden states from pre-trained models like BERT, while the CustomModel constructs a classification model based on a pre-trained transformer.

\subsubsection{MeanPooling}
The MeanPooling class aggregates hidden states $\mathbf{H}$ from a BERT-like model using mean pooling:
\begin{equation}
\mathbf{H}_{\text{mean}} = \frac{1}{T} \sum_{t=1}^{T} \mathbf{H}_t
\end{equation}

\subsubsection{CustomModel}
The CustomModel utilizes the hidden states processed by MeanPooling and feeds them into a fully connected layer for classification:
\begin{equation}
\mathbf{y} = \text{softmax}(\mathbf{W} \mathbf{H}_{\text{mean}} + \mathbf{b})
\end{equation}
where $\mathbf{W}$ and $\mathbf{b}$ are learnable parameters.

\subsection{Model Fine-tune}
22 models are used for integration, and each model learns embedding representations of different dimensions.

\subsubsection{DeBERTa}
DeBERTa (Decoding-enhanced BERT with Disentangled Attention) enhances BERT by introducing disentangled attention mechanisms and a decoding layer. The disentangled attention mechanism separates the absolute and relative positions of words, improving the model's ability to capture syntactic and semantic information:
\begin{equation}
\mathbf{H}_{\text{DeBERTa}} = \text{DisentangledAttention}(\mathbf{H})
\end{equation}
where $\mathbf{H}$ represents the hidden states of the input sequence.

\subsubsection{ALBERT}
ALBERT (A Lite BERT) reduces the number of parameters by factorizing the embedding parameterization and sharing parameters across layers. This makes ALBERT computationally efficient while maintaining performance:
\begin{equation}
\mathbf{H}_{\text{ALBERT}} = \text{SharedLayerNorm}( \text{FactorizedEmbedding}(\mathbf{X}) )
\end{equation}
where $\mathbf{X}$ is the input sequence.

\subsubsection{BART}
BART (Bidirectional and Auto-Regressive Transformers) combines the strengths of BERT and GPT by using a bidirectional encoder and an auto-regressive decoder. This model is effective for text generation and classification tasks:
\begin{equation}
\mathbf{H}_{\text{BART}} = \text{Decoder}(\text{Encoder}(\mathbf{X}))
\end{equation}

\subsubsection{ELECTRA}
ELECTRA (Efficiently Learning an Encoder that Classifies Token Replacements Accurately) introduces a novel pre-training task that involves replacing tokens in the input with plausible alternatives and training the model to distinguish between original and replaced tokens:
\begin{equation}
\mathcal{L}_{\text{ELECTRA}} = -\sum_{t=1}^{T} [y_t \log(p_t) + (1 - y_t) \log(1 - p_t)]
\end{equation}
where $y_t$ is the true token and $p_t$ is the predicted probability of the token being original.

\subsubsection{GPT-2}
GPT-2 (Generative Pre-trained Transformer 2) is an auto-regressive language model that generates coherent and contextually relevant text by predicting the next word in a sequence:
\begin{equation}
P(X) = \prod_{t=1}^{T} P(x_t | x_{1:t-1})
\end{equation}
where $X$ is the input sequence and $x_t$ is the token at position $t$.

\subsubsection{T5}
T5 (Text-to-Text Transfer Transformer) frames all NLP tasks as text-to-text problems, allowing for a unified approach to various tasks such as translation, summarization, and classification:
\begin{equation}
\mathbf{H}_{\text{T5}} = \text{Decoder}(\text{Encoder}(\mathbf{X}))
\end{equation}
T5 employs a sequence-to-sequence architecture where both input and output are treated as text strings.
Each of these models contributes unique strengths to the ensemble, capturing different aspects of the data to improve overall performance.

\subsection{Pseudo-label learning}
Mainly sample the Deberta series models for pseudo-label learning, load the model to initialize the weights. Each model is divided into two modes:
\begin{itemize}
    \item \ Pre-train with pseudo-labels and then fine-tune with only the given training data.
    \item \ Connect the pseudo-labels with the given training data and train all of these data.
\end{itemize}

\subsection{Model Ensemble}
\begin{figure}[h]
    \centering
    \includegraphics[width=0.5\textwidth]{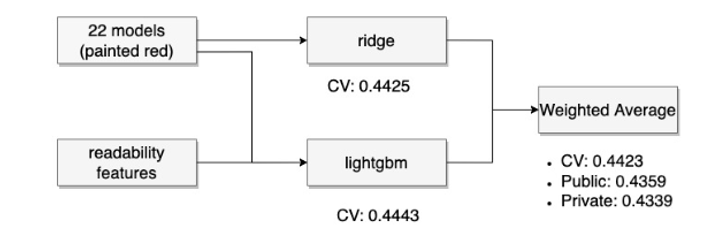}
    \caption{The ensemble pipeline of the model.}
    \label{fig:model_structure}
\end{figure}

Ridge is trained using the predictions of the fine-tuned model as input, while LGB is trained using the predictions and meta-features created by readability. The final output is weighted averaged. The whole model ensemble pipeline is shown in Fig \ref{fig:model_structure}.

\subsection{Data Preprocessing}
Data preprocessing involves stratified k-fold cross-validation and tokenization. The MultilabelStratifiedKFold class ensures balanced label distribution across folds. Missing values in the 'full\_text' column are filled with empty strings to maintain input consistency:
\begin{equation}
\text{full\_text}_i = \text{fillna}(\text{full\_text}_i, "")
\end{equation}
A pre-trained tokenizer, specified by the configuration variable \texttt{CFG.model}, tokenizes the text, preparing it for model input.

\subsection{Loss Function}
The primary loss function used is the Binary Cross-Entropy (BCE) loss, which is suitable for multi-label classification tasks:
\begin{equation}
\mathcal{L}_{\text{BCE}} = -\frac{1}{N} \sum_{i=1}^{N} \left[ y_i \log(p_i) + (1 - y_i) \log(1 - p_i) \right]
\end{equation}
where $y_i$ is the true label and $p_i$ is the predicted probability.

\section{Evaluation Metric}
We utilize several metrics to evaluate model performance, including Root Mean Squared Error (RMSE) and F1-score. The RMSE is defined as:
\begin{equation}
\text{RMSE} = \sqrt{\frac{1}{N} \sum_{i=1}^{N} (y_i - \hat{y}_i)^2}
\end{equation}
where $y_i$ and $\hat{y}_i$ are the true and predicted labels, respectively. The F1-score, a harmonic mean of precision and recall, is given by:
\begin{equation}
\text{F1-score} = 2 \cdot \frac{\text{precision} \cdot \text{recall}}{\text{precision} + \text{recall}}
\end{equation}

\section{Experimental Results}
The models were evaluated on a public and private test set, with performance measured by the metric we mentioned before. The results are summarizedin Table~\ref{tab:performance_metrics}:

\begin{table}[h!]
\centering
\caption{Performance Metrics}
\label{tab:performance_metrics}
\begin{tabular}{|c|c|c|}
\hline
\textbf{Model} & \textbf{RMSE} & \textbf{F1-score} \\
\hline
deberta + lr & 0.423 & 0.561  \\
\hline
deberta + SVR & 0.401 & 0.672 \\
\hline
deberta + GPT2  + lr & 0.392 & 0.741 \\
\hline
deberta + GPT2 + ridge & 0.354 & 0.782 \\
\hline
deberta + roberta + t5 + gpt + lgbm/ridge & \textbf{0.321} & \textbf{0.804} \\
\hline
\end{tabular}
\end{table}

\section{Conclusion}
This study demonstrates the efficacy of combining multiple pre-trained models with pseudo-labeling and ensemble techniques for multi-label text classification. Our approach significantly enhances performance metrics, showcasing the potential for future improvements in the domain of machine learning and deep learning.

 \bibliographystyle{IEEEtran}
    \bibliography{references}

\end{document}